\documentclass[journal]{IEEEtran}
\usepackage{graphicx}
\usepackage{amsmath}
\usepackage{amssymb}
\usepackage{booktabs}
\usepackage[sort, numbers]{natbib}
\usepackage{amssymb}
\usepackage[utf8]{inputenc} % allow utf-8 input
\usepackage[T1]{fontenc}    % use 8-bit T1 fonts
\usepackage{hyperref}       % hyperlinks
\usepackage{url}            % simple URL typesetting
\usepackage{booktabs}       % professional-quality tables
\usepackage{amsfonts}       % blackboard math symbols
\usepackage{nicefrac}       % compact symbols for 1/2, etc.
\usepackage{microtype}      % microtypography
\usepackage[table]{xcolor}
\usepackage{tikz}
\usepackage{xcolor}         % colors
\usepackage{tabularx}
\usepackage{makecell}
\usepackage{utfsym}
\usepackage{stfloats}
\usepackage{orcidlink}

\usepackage{bm}
\usepackage{url}            % simple URL typesetting
\usepackage{amsfonts,amsmath}       % blackboard math symbols
\usepackage{nicefrac}       % compact symbols for 1/2, etc.
\usepackage{microtype}      % microtypography
\usepackage{xcolor}
\usepackage{tabu}
\usepackage{multirow}       % multiple rows in table
\usepackage{graphicx}
\usepackage{graphics}
\usepackage{tabularx}
\usepackage{makecell}
\usepackage{caption}
\usepackage{wrapfig}
\usepackage{enumitem}
\usepackage{subcaption}
\usepackage{algorithm}
\usepackage{arydshln}
\usepackage{algpseudocode}
\usepackage{color, colortbl}

\usepackage{listings}
\definecolor{Gray}{gray}{0.95}
\definecolor{orange}{rgb}{0.9,0.5,0}

\usepackage{pifont}% http://ctan.org/pkg/pifont
\usepackage[capitalize]{cleveref}
\crefname{section}{Sec.}{Secs.}
\Crefname{section}{Section}{Sections}
\Crefname{table}{Table}{Tables}
\crefname{table}{Tab.}{Tabs.}
\begin{document}
	\title{Frequency-Integrated Transformer for Arbitrary-Scale Super-Resolution}
	%\author{Xufei Wang, Fei Ge, Ling Zheng, Shizhuang Weng
	%\thanks{Corresponding author: Ling Zheng, Shizhuang Weng, Email: zhengling@ahu.edu.cn, weng\_1989@126.com,
	%Organization: Anhui University, No. 111, Jiulong Road, Hefei Economic and Technological Development Zone,  Anhui,hefei, 230601,  China.}}
    \author{Xufei Wang~\orcidlink{0009-0006-8442-9526}, Fei Ge~\orcidlink{0009-0009-1187-9691}, Jinchen Zhu, Mingjian Zhang, Qi Wu, Jifeng Ren Shizhuang Weng~\orcidlink{0000-0002-7147-8496}
    \thanks{Corresponding author:  Shizhuang Weng, Email:  weng\_1989@126.com
    Organization: Anhui University, No. 111, Jiulong Road, Hefei Economic and Technological Development Zone,  Anhui,hefei, 230601,  China.}}

	\maketitle
	
	\begin{abstract}Methods based on implicit neural representation have demonstrated remarkable capabilities in arbitrary-scale super-resolution (ASSR) tasks, but they neglect the potential value of the frequency domain, leading to sub-optimal performance. We proposes a novel network called Frequency-Integrated Transformer (FIT) to incorporate and utilize frequency information to enhance ASSR performance.
	FIT employs Frequency Incorporation Module (FIM) to introduce frequency information in a lossless manner and Frequency Utilization Self-Attention module (FUSAM) to efficiently leverage frequency information by exploiting spatial-frequency interrelationship and global nature of frequency.
	FIM enriches detail characterization by incorporating frequency information through a combination of Fast Fourier Transform (FFT) with real-imaginary mapping. In FUSAM, Interaction Implicit Self-Attention (IISA) achieves cross-domain information synergy by interacting spatial and frequency information in subspace, while Frequency Correlation Self-attention (FCSA) captures the global context by computing correlation in frequency.
	Experimental results demonstrate FIT yields superior performance compared to existing methods across multiple benchmark datasets.
	%FIM obtains the clearest visual feature maps among existing methods by enriching detail characterization. Frequency error map (FEM) proved that IISA achieves the slightest frequency error by interacting information in a multi-subspace. Local attribution map (LAM) demonstrates the effectiveness of FCSA in catching the global context by computing frequency correlation.
	Visual feature map proves the superiority of FIM in enriching detail characterization. Frequency error map validates IISA productively improve the frequency fidelity. Local attribution map validates FCSA effectively captures global context.  
	%Future research analyzes the dynamic change pattern of frequency information, adaptively assigns band weights to optimize computational efficiency, and enhances spatial correlation by combining position coding to further deepen the application and exploration of frequency information in super-resolution tasks.

	\end{abstract}
	\begin{IEEEkeywords}
		Super-resolution, Arbitrary-Scale, Frequency, Transformer.
	\end{IEEEkeywords}

	\section{Introduction}
\label{sec:intro}
Single image Super-Resolution (SISR), a process that specializes in reconstructing high-resolution (HR) images from low-resolution (LR) images has been widely used in satellite probing, medical screening and security monitoring \cite{srcnn,swinir,emt}.The emergence of Convolutional Neural Networks (CNNs) and Transformer\cite{srcnn,metalearning,blindsr,blindsr1,camixer} architectures has ushered in a transformative era for Single Image Super-Resolution (SISR). Pioneering CNN- and Transformer-based methodologies, including EDSR \cite{edsr}, RDN \cite{rdn}, and SwinIR \cite{swinir}, have achieved notable breakthroughs in reconstruction accuracy and perceptual quality, establishing new benchmarks for the field.
Due to the fact that real-world enhancement of LR images to non-integer and non-fixed scales is required, Arbitrary-Scale Super-Resolution (ASSR) has been a surge of interest among researchers in recent years \cite{metasr,liif,lte,clit,srno,ciaosr,amiassr}. 
Chen et al. \cite{liif} introduced implicit neural representation (INR) for arbitrary-scale super-resolution (ASSR)to generate RGB values by mapping encoder-extracted LR image features to HR coordinates through a continuous function that leverages spatial distance relationship, achieving favorable results. Recent researchers have improved ASSR performance by enriching information.
%Xu et al. and Liu et al. extended the positional encoding of INR, and Lee et al. added unprecedented texture information to INR.
Li et al. \cite{lte} optimized the expression of INR by adding texture information
Wei et al. \cite{srno} introduced the mapping of different image pairs in finite dimensions.
%Zhao et al. \cite{amiassr} extract different scale feature information by making adaptive changes in the size and shape of the convolution kernel.
Moreover, Cao et al. and Chen et al. \cite{ciaosr,clit} have combined self-attention in Transformer with INR to obtain additional contextual information. Recently, frequency has been increasingly favored by researchers as a unique way of extracting information. Li et al. \cite{craft} used pooling to extract high-frequency details. Kong et al. \cite{fasa} constructed parameter learnable filters to extract critical frequency information. But performance is still limited due to the lossy introduction and inefficient leverage of frequency information.

To address these issues, Chi et al. \cite{ffc} designed a structure combining FFT and convolution to attempt lossless extraction of frequency information.
%Huang et al. \cite{affnet} exploited the excellent semantic adaptation of frequency information to achieve a lightweight large kernel dynamic convolution.
Huang et al. \cite{affnet} proposed to fully utilize the frequency information based on its characteristics instead of introducing it into the network only as extra information. 
Herein, we proposed the Frequency-Integrated Transformer (FIT) to incorporate the frequency information losslessly and utilize the spatial-frequency interrelationship and the global nature of frequency information.
FIT consisting of Frequency Incorporation Module (FIM) and Frequency Utilization Self-Attention module (FUSAM) for ASSR. 
FIM combined FFT and real-imaginary mapping to losslessly incorporate frequency information into the network. 
In FUSAM, IISA realizes cross-domain information synergy by alternately projecting spatial and frequency information into the multi-subspace, 
%for interaction of the two types of information,
FCSA computes correlation in the frequency to leverage the global nature of frequency.
%for enhancing the capability of catching global context. 
Extensive experiments demonstrate our network achieves excellent results on multiple benchmark datasets.

	\section{Related Work}
\label{sec:relate}
\paragraph{Single image super-resolution}
SISR is a low-level visual task that has been utilized to recover from low-resolution (LR) images to high-resolution (HR) images. SRCNN\cite{srcnn} first applied CNN to SISR, marking the entry of SISR into the deep learning era. CNN-based methods are widely utilized, such as EDSR\cite{edsr} and RDN\cite{rdn}. Recently, Transformer-based methods, such as SwinIR\cite{swinir} and SRFormer\cite{srformer}, are becoming popular in SISR due to the fact that they can leverage context information through Self-Attention (SA). But the above methods can only be used for fixed magnification, which limits their deployment in realistic scenarios.

\paragraph{Arbitrary scale super-resolution based on INR}
ASSR is a method capable of improving image resolution at arbitrary scales. INR is a technique for processing continuous signals using a Multilayer Perceptron (MLP) and applied in various visualization tasks, such as object modeling, scene reconstruction and structure rendering \cite{3dscene_0,3dscene_1,3dscene_2,3dscene_3}. Chen et al. \cite{liif} firstly uses INR for predicting RGB values by the feature around the LR coordinate corresponding to the HR coordinates in continuous domain to perform ASSR. 
%Xu et al. fused coordinate coding of INR and periodic coding. Liu et al. enrich position encoding by aggregating diverse information in the pixel area. Yang et al. applies point-to-point INR for RGB value prediction and clusters adjacent pixels to the center pixel point thus obtaining thin and sharp edges. .
Lee et al. \cite{lte}presented the Local Texture Estimator to add new texture information, which effectively improves the representation of INR.
Wei et al. \cite{srno} regraded the mapping between LR-HR image pairs as a continuous function and fitted a common latent basis for such functions using INR modified by the Galyokin attention mechanism.
%Fu et al. Add spatial texture information to the model to improve the robustness of the model against image degradation.
Moreover, Cao et al. \cite{ciaosr} and Chen et al. \cite{clit} recognised the importance of contextual information for ASSR, and they combined the Transformer with INR approach to embed contextual information into the network.
Zhao et al. \cite{amiassr} extract different scale feature information by making adaptive changes in the size and shape of the convolution kernel.
%But these only focus on spatial information to achieve limited performance improvement.
But these only focus on spatial information leads to undesirable outcomes.
\paragraph{Frequency domain operations in Image Enhancement}
Frequency domain operations are important in conventional signal processing fields \cite{f2000}. Some researchers first introduced the frequency domain operations to deep learning as a kind of tool to measure the validity \cite{robustness} and generalization \cite{generalization1} of models. Since frequency information contains unique characteristics distinct from spatial information \cite{f2015}, some researchers have proposed to use frequency information to boost model performance. 
Li et al. \cite{craft} used pooling to discard low-frequency information to obtain high-frequency details. Kong et al. \cite{fasa} advocate the use of parameter learnable filters to discriminatively retain critical frequency information only. Chi et al. \cite{ffc} preprocessed complex-valued frequency inputs to adequately extract information through convolution. Wang et al. \cite{sfmnet} enriched the detailed characterization by extracting the amplitude and phase components of the frequency information. Huang et al. \cite{affnet} constructed adaptive frequency filtering token mixer for implementing lightweight large-kernel dynamic convolution based on the excellent semantic adaptation of frequency information to fully leverage the frequency information according to the its characteristics.
Lossless introduction and efficient utilization of frequency information is crucial for ASSR improvement.

%Therefore, we should construct a network capable of introducing frequency information losslessly through FFT and convolution while fully utilize it according to the frequency characteristics.
%some researchers \cite{esrt,iformer,craft,afcnet} have proposed the pooling methods to obtain high-frequency details by discarding low-frequency information. Subsequently, some researchers \cite{gfnet,spectformer,affnet,fasa} recognising the importance of low-frequency information have suggested the use of filters and nonlinear activation functions to retain critical frequency information. 
%Further, recent researchers \cite{sfmnet,fadformer} have proposed methods based on Fast Fourier Transform FFT and information preprocessing to extract the frequency information as a optimal manner, but their processing leads to still imperfect extracted frequency information. 
%The above methods lose the original information, so the extracted frequency information is incomplete and the unique characteristics of the frequency information are unrecognized, resulting in ineffective use of the frequency information. This leads to the fact that they can only get limited performance improvement from the frequency information.

\begin{figure*}[t]
	\centering
	\begin{subfigure}{1\linewidth}
		\centering
		\includegraphics[width=1\linewidth]{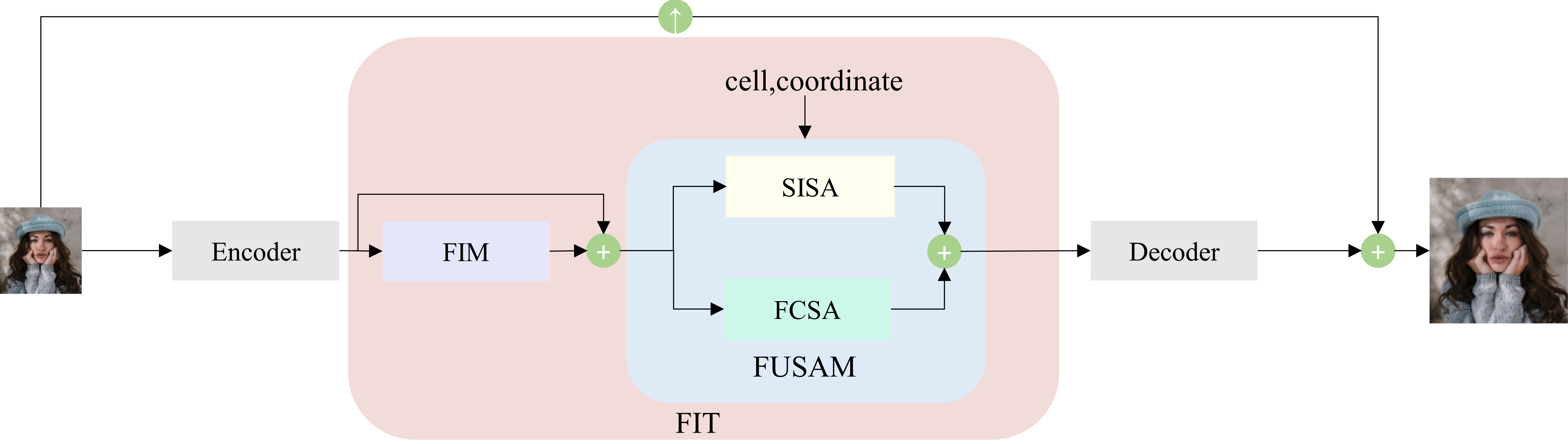}
	\end{subfigure}
	\caption{Overall architecture for ASSR}
	\label{overall}%文中引用该图片代号
\end{figure*}

\begin{figure*}[tb]
	\centering
	\begin{subfigure}{1\linewidth}
		\centering
		\includegraphics[width=1\linewidth]{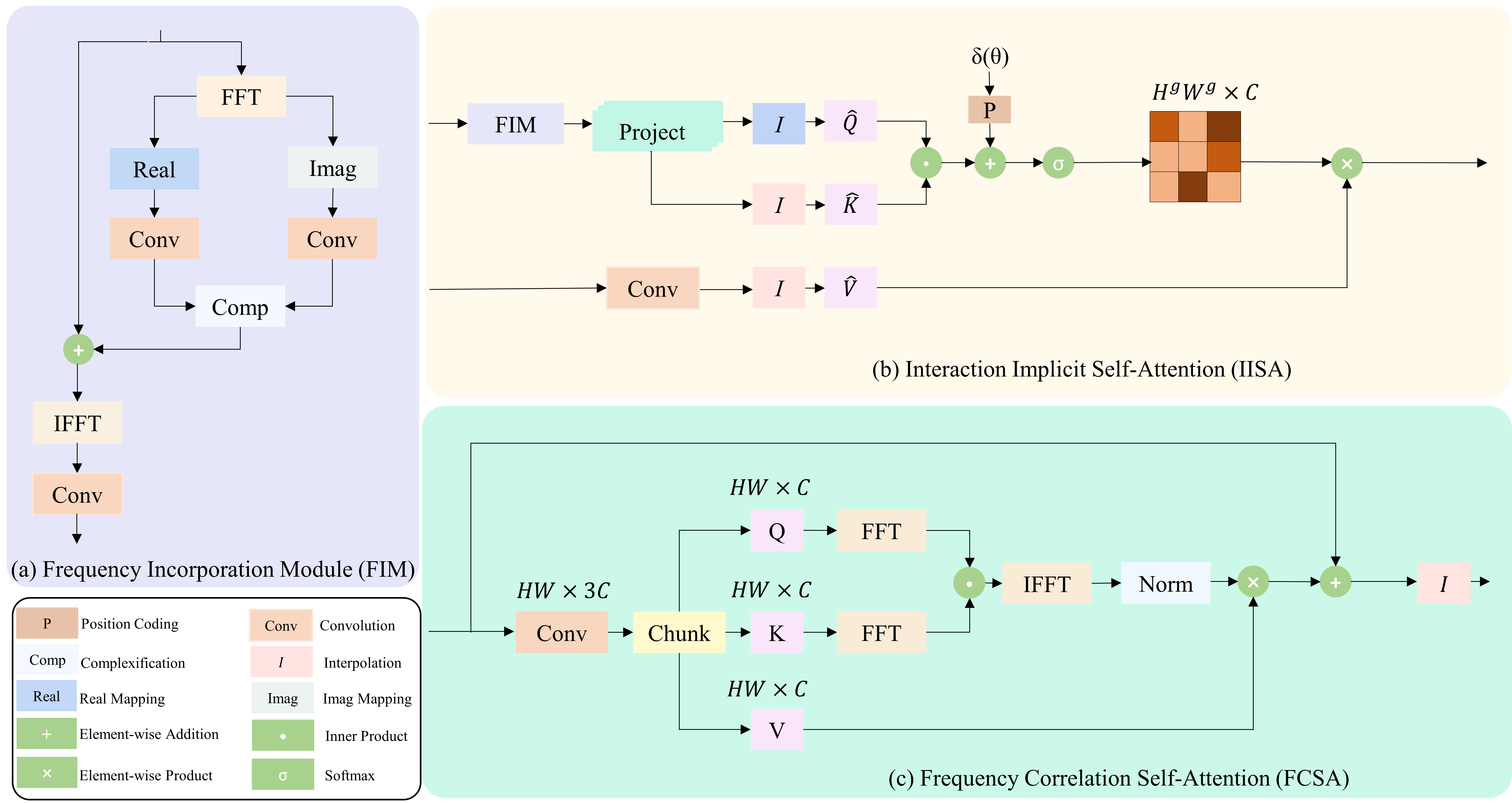}
	\end{subfigure}
	\caption{(a) Overview of FIM structure
		, we use real-imaginary mapping and convolution to extract the frequency information and use element-wise addition and point-wise convolution to incorporate the frequency information into the network. (b) Overview of IISA structure, we project the input into multi-subspace to enable initial interaction with information from different domains and re-interaction using multi-head attention. (c) Overview of IISA structure, we converts $Q$ and $K$ to the frequency domain for computing correlation.}
	\label{overview}%文中引用该图片代号
	
\end{figure*}
\begin{figure*}[t]
	\centering
	\begin{subfigure}{1\linewidth}
		\centering
		\includegraphics[width=1\linewidth]{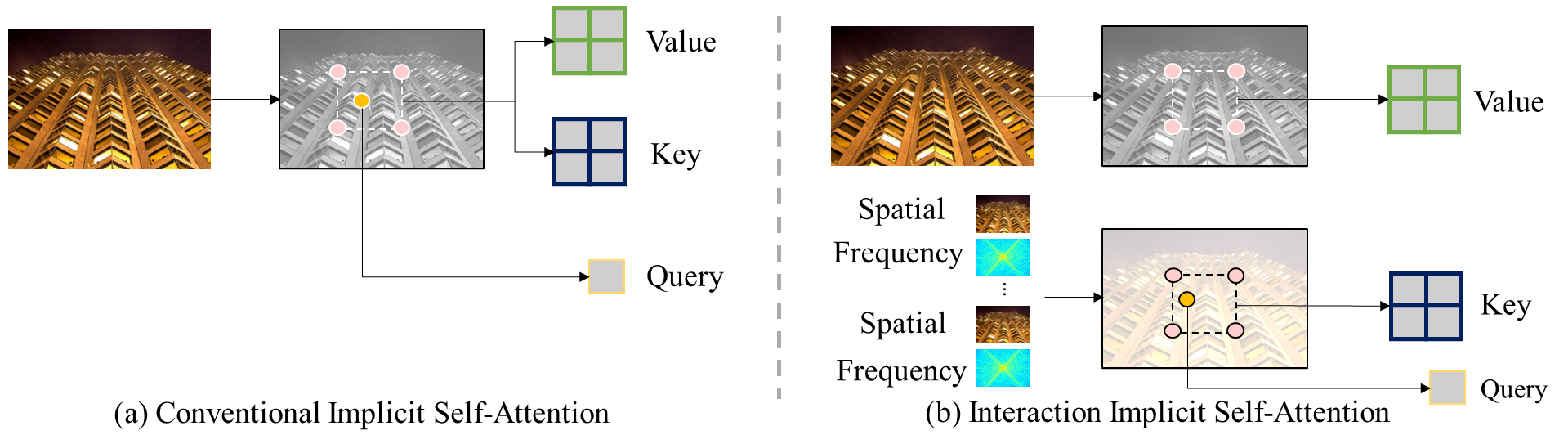}
	\end{subfigure}
	\caption{Overview of differences between conventional Implicit Self-Attention and Interaction Implicit Self-Attention. We alternately project spatial and frequency information into the multi-subspace for cross-domain information interaction to obtain query and key, where the green image is a pseudo-colored frequency map representing the frequency information}
	\label{isa}%文中引用该图片代号	
\end{figure*}
%\begin{figure*}[h]
	%\centering
	%\begin{subfigure}{1\linewidth}
		%\centering
		%\includegraphics[width=1\linewidth]{qualitative_img}
	%\end{subfigure}
	%\caption{Qualitative comparison of MSIT with using RDN as the encoder.}
	%\label{qualitative_0}%文中引用该图片代号
%\end{figure*}

%\begin{figure*}[t]
	%\centering
	%\begin{subfigure}{0.8\linewidth}
		%\centering
		%\includegraphics[width=1\linewidth]{qualitative_1}
	%\end{subfigure}
	%\caption{Qualitative comparison with stepwise incremental SR of MSIT with using RDN as the encoder.}
	%\label{qualitative_1}%文中引用该图片代号
%\end{figure*}

\section{Methodology}
\label{sec:meth}
In this section, we describes the framework and main modules of our network.
\subsection{Overall Pipeline of Framework}
Our network in Fig. \ref{overall} consists of Encoder, FIT and Decoder. The network is capable of generating the HR image $I_{HR}\in\mathbb{R}^{\eta_hH \times \eta_wW \times 3}$ at HR coordinates $\theta^h\in \{\theta^h_j \}^{j=G_h}_{j=1}$ from the given LR image $I_{LR}\in\mathbb{R}^{H \times W \times 3}$ at LR coordinates $\theta^l\in \{\theta^l_j\}^{j=G_l}_{j=1}$ in the arbitrary amplified scale $\eta:\{\eta_h, \eta_w\}$, where $G_h$ and $G_l$ denote HR and LR coordinate of the 2D space in continuous domain. The encoder $E_\psi$ first extracts the spatial feature $\mathcal {Z}_{in}\in\mathbb{R}^{H \times W \times C}$ from $I_{LR}$. Next, $\mathcal Z_{in}$ is delivered into FIM for extracting frequency information to generate $\mathcal{Z}_{FIM}$. And we use the jump connection to link $\mathcal Z_{in}$ and $\mathcal{Z}_{FIM}$. Then FUSAM use $\mathcal{Z}_{FIM}$ to obtain attention feature $\mathcal{Z}_{FUSAM}$ with $\theta^h$ and $Cell$, where $Cell$ signifies the form of the query pixel. Finally, we use bilinear interpolation to amplify the $I_{LR}$ to $I^{\uparrow}_{LR}\in \mathbb{R}^{\eta_hH \times \eta_wW \times 3}$ and merge it with output of decoder $D_\psi$ through element-wise addition to produce $I_{HR}\in\mathbb{R}^{\eta_hH \times \eta_wW \times 3}$ pixel by pixel. The entire process can be formulated as follows:
%\begin{equation}
%I_{HR} = \bold {D_{\psi}}({  \bold {FUSAM}((\bold {FIM}( \bold{E_{\psi}}(I_{LR}))+\bold{E_{\psi}}(I_{LR})), Cell, \theta_h )})+ I^{\uparrow}_{LR}
%\end{equation}
%\begin{equation}
%I_{HR} = \bold {D_{\psi}}({\mathcal{Z}_{out}}) + I^{\uparrow}_{LR}
%\end{equation}
\begin{equation}
I_{HR} = \bold {D_{\psi}}({ \bold {FIT}( \bold{E_{\psi}}(I_{LR}), Cell, \theta_h )})+ I^{\uparrow}_{LR}
\end{equation}
\subsection{Frequency Incorporation Module}
Fourier transform is an important tool for processing image signals \cite{f2000},  
but the complex-valued frequency information obtained from FFT is unable to be combined with convolution, resulting in unavoidable loss of information.
Gao et al. \cite{fadformer} proposed collapsing the complex-valued frequency information into the channel dimension. Wang et al. \cite{sfmnet} transformed the frequency information to a polar coordinate system and extracted the amplitude and phase component of the frequency. 
%But some numbers of the tensor representing the details of the image will be incorrectly processed to non-number, such as nan, posinf and neginf during the process, resulting in the loss of some potential feature of detailed texture. 
%but this method performs feature extraction while destroying the channel dependence.
%Since the convolution operation adheres to the distributive property, where the summation of convolutions applied to individual components equals the convolution applied to the completion.
%Thus the sum of convolution of each part of complex-valued frequency information to be equivalent to the convolution of whole. 
%Fig. \ref{overview} (a) shows our Frequency Incorporation Module (FIM) to extract the frequency information in a lossless manner by combining FFT and real-imaginary mapping.
The convolution operation adheres to the distributive property, which means that summation of convolutions applied to individual components equals the convolution applied to the entire input.
%We design the Frequency Incorporation Module (FIM) in Fig. \ref{overview} (a) to extract the  complex-valued frequency information in a lossless manner by combining FFT and real-imaginary mapping. 

We design the Frequency Incorporation Module (FIM) in Fig. \ref{overview} (a) to extract the  complex-valued frequency information by combining FFT and real-imaginary mapping. This is a lossless manner due to the real-imaginary mapping is constant mapping.

In FIM, we perform further information extraction in the spatial and frequency domain for $\mathcal{Z}_{in}$. We used FFT to convert $\mathcal{Z}_{in}$ to complex tensor $\mathcal{Z}_{FFT}$ and map it into real information $\mathcal{Z}_{real}$ with imaginary information $\mathcal{Z}_{imag}$ to perform frequency information extraction. Then we complexification the $\mathcal{Z}_{real}$ and $\mathcal{Z}_{imag}$ into $\mathcal{Z}_{FFT}^{\prime}$. The whole process in frequency can be described as:
\begin{equation}
\mathcal{Z}_{FFT}^{\prime} = Comp({Conv}(\mathcal{Z}_{real}), {Conv}(\mathcal{Z}_{imag}))
\end{equation}
Finally, we combine $\mathcal{Z}_{FFT}^{\prime}$ and $\mathcal{Z}_{in}$ by skip connection with further using Inverse Fast Fourier Transform (IFFT) and point-wise convolution for modulation to incorporate $\mathcal{Z}_{FFT}^{\prime}$ into the network:
\begin{equation}
\mathcal{Z}_{FIM} = {PConv}(\mathcal {F}^{-1}(\mathcal{Z}_{FFT}^{\prime} + (\mathcal{Z}_{in})))
\end{equation}
Where the $Comp$ denotes the the $\mathcal {F}$ denotes the FFT and the $\mathcal {F}^{-1}$ denotes the IFFT.

\subsection {Frequency Utilization Self-Attention module}
\subsubsection{Rethinking the Fourier Transform}
The process of applying the Fourier transform to a single-channel image $f(x,y)$ can be expressed as follows:
\begin{equation}
	\mathcal{F}(u, v) = \frac{1}{\sqrt{MN}} \sum_{x=0}^{M-1} \sum_{y=0}^{N-1} f(x, y) \cdot e^{-i 2\pi \left( \frac{ux}{M} + \frac{vy}{N} \right)} \label{fft}
\end{equation}
Where $(x,y)$ represents the spatial coordinates of the image, the $\mathcal{F}(u, v)$ is a complex value in the frequency domain, representing the frequency component, the $u=0,1,\dots,M-1$ and $v=0,1,\dots,N-1$ are the coordinates in the frequency domain, the $M$ and $N$ represent the width and height of the image. We can learn that every value in $\mathcal{F}(u,v)$ is the aggregation of all values in $f(x,y)$.
%from Eq. \ref{fft} 
Thus the spatial-frequency interrelationship and the global nature of frequency is critical for utilization of frequency information.

We designed Frequency-utilization Self-Attention module (FUSAM) containing Interaction Implicit Self-Attention (IISA) and Frequency Correlation Self-Attention (FCSA), as shown in Fig. \ref{overall}.
%this dual self-attention structure ensures that each self-attention makes full use of spatial-frequency interrelationship and global nature respectively and achieves complementary enhancement through element-wise addition. 
%IISA achieves cross-domain synergy of information by projecting spatial and frequency information to multi-subspace. FCSA efficiently catches global context by computing correlation in the frequency domain. 
IISA and FCSA are connected through element-wise addition:
\begin{equation}
\mathcal {Z}^{FUSAM} = \mathcal {Z}^{IISA}+\mathcal  {Z}^{FCSA}
\end{equation}
%IISA computes pixel self-attention $\mathcal {Z}^{Attn}_{IISA}$ using spatial domain-based position coding and $Cell$ in the spatial domain, while FCSA compute attention potential feature $\mathcal {Z}^{Attn}_{FCSA}$ in the frequency domain for providing diffrernt attention potential features from pixel self-attention and widening the receptive field of FUSAM. 

\subsubsection{Interaction Implicit Self-Attention}
%Current methods for interacting spatial and frequency information fail to take the interrelationship between the two types of information into account, resulting in adequate synergy of cross-domain information.

Current methods for interacting spatial and frequency information neglect interrelationship between the two types of information.
Chen et al. \cite{clit} enhancing information interaction in Implicit Self-Attention (ISA) through Multi-Head Self-Attention (MHSA).
And the analogous mechanisms can be generalized to cross-domain information synergy. 
We proposed Interaction Implicit Self-Attention (IISA) to achieve cross-domain synergy of information by projecting spatial and frequency information to multi-subspace.
%Fig. \ref{overview} (b) illustrates Interaction Implicit Self-Attention (IISA) we proposed for cross-domain synergy of information by projecting spatial and frequency information into multi-subspace.

The framework of IISA is shown in Fig. \ref{overview}, IISA uses FIM to further extract frequency information from the input $\mathcal {Z}_{FIM}$ to obtain $\mathcal {Z}_{FIM}^{\prime}$. Then, the matrices project $\mathcal {Z}_{FIM}^{\prime}$ into several different subspaces. The matrices alternately project spatial and frequency information into different subspaces and use linear layers for fusing different subspaces into $Q$ to enable initial interaction of space and frequency information as shown in Fig. \ref{isa}. The whole process can be expressed as:
\begin{equation}
	Q=Linear\left\{
	\begin{array}{rcl}
		Q_s &=  \mathcal {Z}_{FIM}^{\prime} \times \mathcal {W}_n& n=1, 3, ...2i-1\\
		Q_f &=  \mathcal {Z}_{FIM}^{\prime} \times \mathcal {W}_m& m=2, 4, ...2i
	\end{array} \right.
\end{equation}
where $\mathcal {W}_n$ represents the spatial matrix and $\mathcal {W}_m$ represents the frequency matrix. The $i$ represents the half of all projection matrices.

%We use the above method to obtain $Q$ while 
The generation of $V$ follow the multi-head attention. IISA calculates correlation by sampling the queried grid $\hat {\theta}^l = \{\hat {\theta}^l_j\}^{j = {H_g}{W_g}}_{j=1} $, where $H_g$ and $W_g$ indicate the height and width of $\hat {\theta}^l$. And center coordinate of $\hat {\theta}^l$ is the LR coordinate closest to the queried HR coordinate $\hat {\theta}^h$. The query vector $\hat Q \in\mathbb {R}^{1 \times C}$ at HR coordinate $\hat {\theta}^h$ is obtained by using bilinear interpolation from $Q$, while the $Q$ is interpolated to be key vector $\hat K \in\mathbb {R}^{{H_g}{W_g} \times C}$ at LR queried grid $\hat {\theta}^l$. Then the value vector  $\hat V \in\mathbb {R}^{{H_g}{W_g} \times C}$ at LR queried grid $\hat {\theta}^l$ is obtained by using neighborhood interpolation from $V$. We use MHSA to achieve information re-interaction:
\begin{equation}
\mathcal {Z}^{IISA} = Concat (Softmax(\mathcal{F}(\delta(\theta))_\mu + \cfrac{\hat Q_\mu\hat K^{T}_\mu}{G} )\times \hat V_\mu)
\end{equation}
%\mathcal{B}(\delta(\theta))
\begin{equation}
\begin{split}
\mathcal{F}(\delta(\theta)) = &\mathcal{F}[sin(\phi_1\delta(\theta)), cos(\phi_1\delta(\theta)), \\
&..., sin(\phi_p\delta(\theta)), cos(\phi_p\delta(\theta))]
\end{split}
\end{equation}
\begin{equation}
G = \sqrt{d_k/H}
\end{equation}
\begin{equation}
\delta(\theta)= \hat\theta^h- \hat \theta^l_j
\end{equation}
where $\mathcal{F}$ stands for the fully connected layer consisting of linear units, $\delta$ represents sinusoidal position encoding. $G$ is the hyperparameter for the size of the matrix that generates the multi-head. The $d_k$ and $H$ represent channel dimension of vector $\hat K$ and the number of attention heads, respectively. The $\mu \in [1,2,...,H]$ denotes every attention head. The hyperparameter $p$ is set to 10 while $H$ is set to 8 in our work.

\subsubsection{Frequency Correlation Self-Attention}
Existing method is inefficient in leveraging the global nature of frequency information. 
Recently, Cui et al. \cite{OKnet} introduced the global nature of frequency information into the network by performing element-wise product between the input and frequency information as global weight. SA proposed by Vaswani et al. inherently excels in catching global context \cite{attention}. 
We designed Frequency Correlation Self-Attention (FCSA) to utilize frequency correlation as the attention weight, thereby efficiently capturing global context.
%some researchers \cite{sfmnet,fasa} claims that the capture of global context can be useful in enhancing the ability of modules to model long-distance dependency.
%Therefore, We computing frequency correlation can utilize the global nature of frequency to capture the receptive field in size of the image for capturing global context more efficiently.

The structure of FCSA is shown in Fig. \ref{overview} (c), FCSA will first process the  $\mathcal {Z}_{FIM}\in\mathbb{R}^{H \times W \times C}$ to $\mathcal {\tilde{Z}}_{FIM}\in\mathbb{R}^{H \times W \times 3C}$. Then, the $\mathcal {\tilde{Z}}_{FIM}\in\mathbb{R}^{H \times W \times 3C}$ will be chunked into $ \tilde Q\in\mathbb{R}^{H \times W \times C}$, $\tilde K\in\mathbb{R}^{H \times W \times C}$ and $\tilde V\in\mathbb{R}^{H \times W \times C}$ in the channel dimension. Then, we use the FFT for converting $ \tilde Q$ and $ \tilde K$ to the frequency and compute the correlation with the following formula:
\begin{equation}
F_{attn}= {Norm}(\cfrac{\mathcal {F}^{-1}((\mathcal {F}(\tilde Q)({\mathcal {F}^T(\tilde K)})))}{\sqrt {d_{\tilde{k}}}})
\end{equation}
where the ${Norm}$ denotes Norm, ${\mathcal{F}^T}$ is transpose of the FFT. Then attention can be calculated by:
\begin{equation}
attn=F_{attn} \times \tilde V\
\end{equation}
Then we added the jump connection for $attn$ to get $Attn$. Finally, we compute the neighborhood interpolation based on $\hat{\theta}^l$ for $Attn$ to get $\widehat {Attn}$ = $\mathcal {Z}^{FCSA}\in\mathbb{R}^{{H_q}{W_q} \times C}$. This ensures that IISA and FCSA compute attention in the same query grid. 
%And addresses the problem that computing frequency-domain attention in attention head of IISA fails to incorporate positional encoding and $Cell$ based on the spatial domain.

	\begin{figure*}[b]
	\centering
	\begin{subfigure}{1\linewidth}
		\centering
		\includegraphics[width=0.9\linewidth]{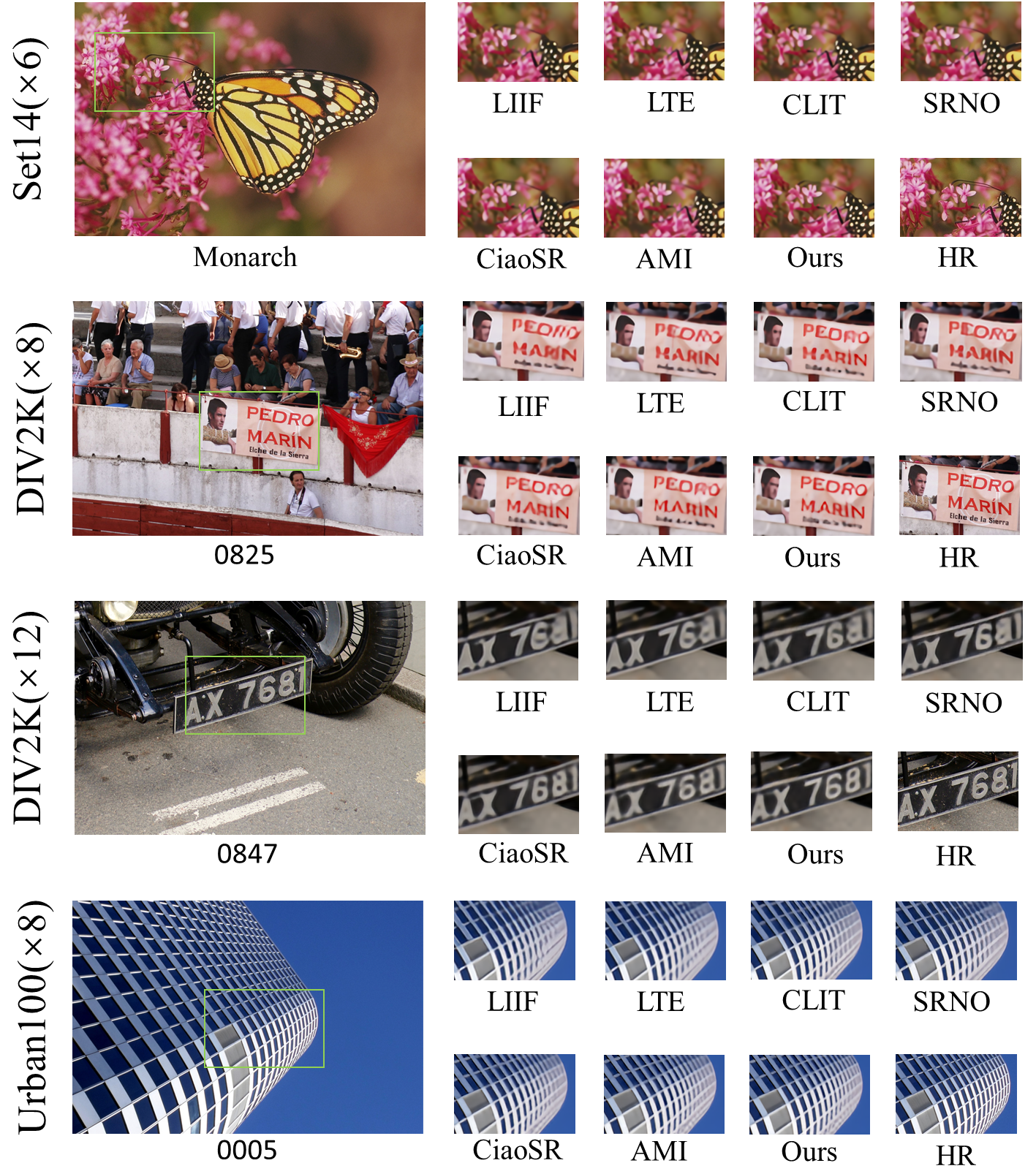}
	\end{subfigure}
	\caption{Qualitative comparison of integer scales with RDN as encoder.}
	\label{integer}%文中引用该图片代号
\end{figure*}

\begin{figure*}[b]
	\centering
	\begin{subfigure}{1\linewidth}
		\centering
		\includegraphics[width=0.9\linewidth]{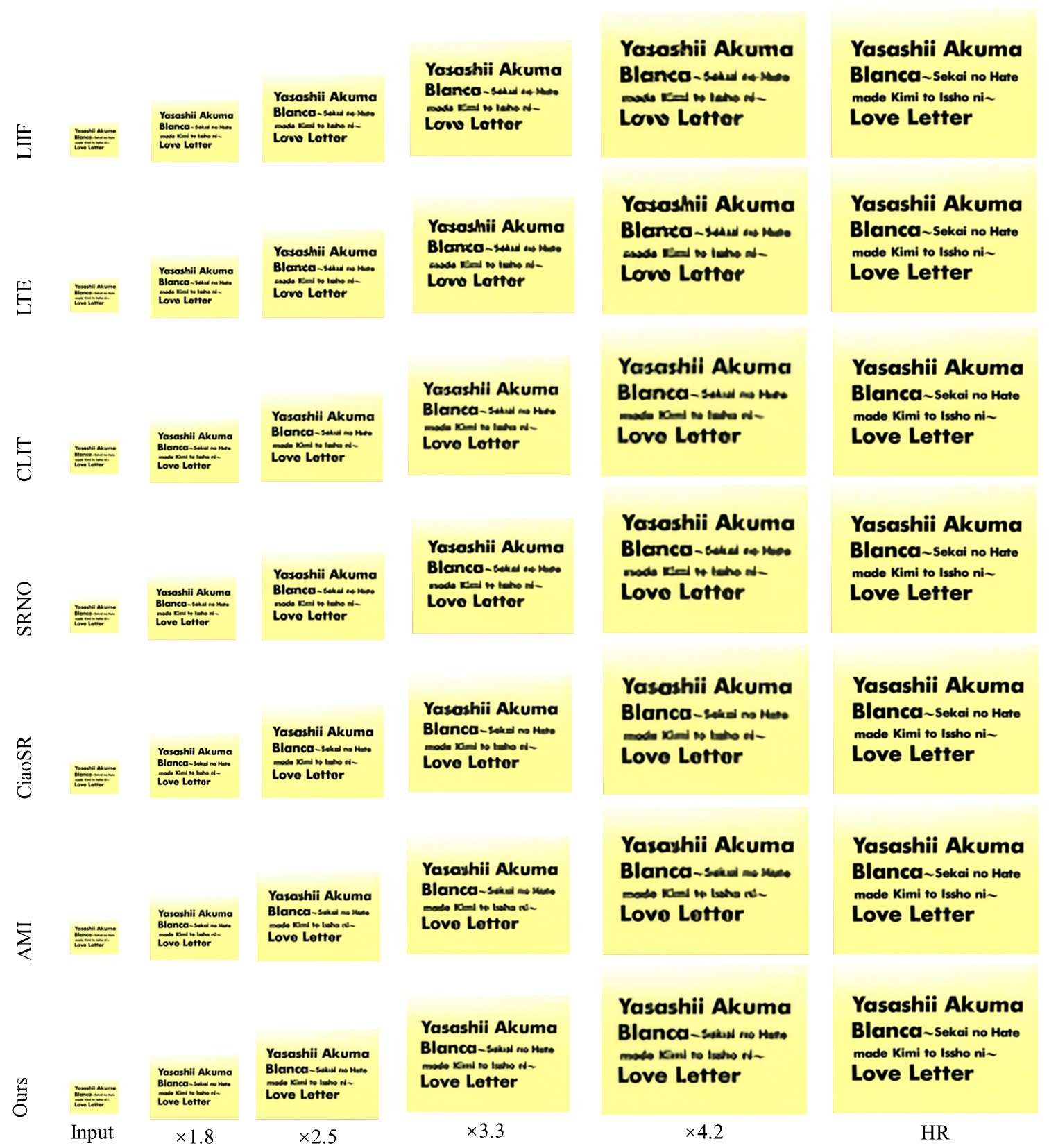}
	\end{subfigure}
	\caption{Qualitative comparison of non-integer scales with RDN as encoder.}
	\label{noninteger}%文中引用该图片代号
\end{figure*}

\begin{figure*}[b]
	\centering
	\begin{subfigure}{1\linewidth}
		\centering
		\includegraphics[width=1\linewidth]{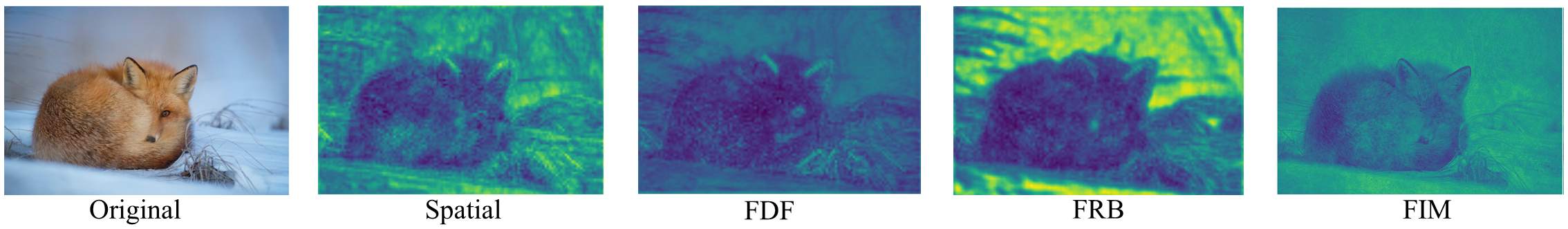}
	\end{subfigure}
	\caption{Results of visual feature map for different frequency information extraction modules.}
	\label{visual}%文中引用该图片代号
\end{figure*}

\section{Experience}
\label{sec:expe}
\begin{table*}[t]
	\tiny
	%\scriptsize
	\centering
	
	\resizebox{0.9\linewidth}{!}{
	\begin{tabu}{lccccccccc}
		\hline
		\specialrule{0em}{1pt}{0pt}
		{Method}  & {$\times2$}  & {$\times3$} &{$\times4$} & {$\times6$}& {$\times12$}& {$\times18$}& {$\times24$}&{$\times30$} \\
		\specialrule{0em}{1pt}{0pt}
		\hline
		\specialrule{0em}{1pt}{0pt}
		\hline
		\specialrule{0em}{1pt}{0pt}
		
		Bicubic \cite{edsr} &31.01 &28.22 &26.66 &24.82 &22.27 &21.00 &20.19 &19.59                   \\
		\specialrule{0em}{1pt}{0pt}
		\hline
		\specialrule{0em}{2pt}{0pt}
		
		EDSR \cite{edsr} &34.55 &30.90 &28.94 &- &- &- &- &-                                             \\
		EDSR-MetaSR \cite{metasr}   &34.64 &30.93 &28.92 &26.61 &23.55 &22.03 &21.06 &20.37         \\
		EDSR-LIIF \cite{liif} &34.67 &30.96 &29.00 &26.75 &23.71 &22.17 &21.18 &20.48                          \\
		%EDSR-UltraSR  &34.69 &31.02 &29.05 &26.81 &23.75 &22.21 &21.21 &20.51             \\
		%EDSR-ITSRN &34.71 &30.95 &29.03 &26.77	&23.71	&22.17	&21.18	&20.49
		%\\
		%EDSR-IPE  &34.72 &31.01 &29.04 &26.79 &23.75 &22.21 &21.22 &20.51                         \\
		EDSR-LTE \cite{lte}  &34.72 &31.02 &29.04 &26.81 &23.78 &22.23 &21.24 &20.53                        \\
		EDSR-CLIT \cite{clit}  &34.82 &31.14 &29.17 &26.93 &23.85 &22.30 &21.27 &20.54                       \\
		EDSR-SRNO \cite{srno} &34.85 &31.11 &29.16 &26.90 &23.84 &22.29 &21.27 &20.56 \\
		EDSR-CiaoSR \cite{ciaosr} &34.91	&31.15	&29.23	&26.95	&23.88	&22.32	&21.32	&20.59
		\\
		EDSR-AMI \cite{amiassr} &34.92	&31.22	&29.23	&26.97	&23.90	&22.34	&21.30	&20.56
		\\
		
		%EDSR-LMI  &34.94	&31.22	&29.25	&27.01	&23.92	&22.36	&21.32	&20.60 
		%\\
		
		EDSR-FIT(Ours) &$\textbf{34.98}$	&$\textbf{31.29}$	&$\textbf{29.32}$	&$\textbf{27.05}$	&$\textbf{23.95}$	&$\textbf{22.39}$	&$\textbf{21.34}$	&$\textbf{20.62}$
		\\
		
		\specialrule{0em}{1pt}{0pt}
		\hline
		\specialrule{0em}{2pt}{0pt}
		RDN \cite{rdn}   &34.94 &31.22 &29.19 &- &- &- &- &-            \\
		RDN-MetaSR \cite{metasr}   &35.00 &31.27 &29.25 &26.88 &23.73 &22.18 &21.17 &20.47             \\
		RDN-LIIF \cite{liif}  &34.99 &31.26 &29.27  &26.99 &23.89 &22.34&21.31 &20.59                            \\
		%RDN-UltraSR   &35.00 &31.30 &29.32 &27.03 &23.73 & 22.36 &21.33 &20.61            \\
		%RDN-ITSRN &35.09	&31.36	&29.38	&27.06	&23.93	&22.36	&21.32	&20.61 \\
		
		%RDN-IPE  &35.04 &31.32 &29.32&27.04 &23.93 &22.38 &21.34  &20.63                                  \\
		RDN-LTE \cite{lte} &35.04 &31.32 &29.33 &27.04 &23.95 &22.40 & 21.36 &20.64                         \\
		RDN-CLIT \cite{clit} &35.10 &31.38 &29.40 &27.12 &24.01 &22.45 &21.38 &20.64                                                \\
		RDN-SRNO \cite{srno} &35.16 &31.42 &29.42 &27.12 &24.03 &22.46 &21.41 &20.68 \\
		RDN-CiaoSR \cite{ciaosr} &35.15	&31.42	&29.45	&27.16	&24.06	&22.48	&21.43	&20.70
		\\
		
		RDN-AMI \cite{amiassr} &35.17	&31.46	&29.46	&27.16	&24.06	&22.49	&21.40	&20.66
		\\
		%RDN-LMI    &35.17	&31.46	&29.49	&27.20	&24.07	&22.50	&21.42	&20.71
		%\\
		RDN-FIT(Ours)     &$\textbf{35.22}$	&$\textbf{31.53}$	&$\textbf{29.56}$	&$\textbf{27.24}$	&$\textbf{24.10}$	&$\textbf{22.53}$	&$\textbf{21.46}$	&$\textbf{20.73}$
		\\
		
		\specialrule{0em}{1pt}{0pt}
		\hline
		
	\end{tabu}
	}
	\caption{Quantitative comparison with the SOTA methods on the DIV2K validation set. 
			The best results are shown in $\textbf{Bold}$.}
	\label{sota_div}
\end{table*}

\subsection{Implementation Details}
\subsubsection{Datasets and Metrics}
We use the training set of DF2K \cite{df2k} to train our network. And we use the validation set of DIV2K \cite{div2k}, Set5 \cite{set5}, Set14 \cite{set14}, Urban100 \cite{U100} and BSD100 \cite{BSD100} as test sets to evaluate our model. 
%We use the Peak Signal-to-Noise Ratio (PSNR), which is widely used in the field of super resolution (SR) , as an evaluation metric for the data to show the superior performance of our model compared to other models.
We adopt the widely used Peak Signal-to-Noise Ratio (PSNR) in image enhancement \cite{tip251,tip252,tip1,tip2,tip3} as the evaluation metric.

\subsubsection{Training Setting}
Our training strategy follows previous research \cite{liif,lte,clit,refconv}. We crop the HR image into image patches of 48$\eta$ $\times$ 48$\eta$, $\eta$ is an amplification factor randomly sampled from the mean distribution of $U(1,4)$. We processed the HR image patches in Pytorch \cite{pytorch} using bilateral interpolation to obtain the corresponding LR image patches. We then using random horizontal flips, vertical flips and 90°rotations for augmenting the LR image patches to enhance the diversity of our dataset. Our ground truth data is obtained by sampling $48^2$ pixels from each HR patch (coordinate-RGB pairs). We used the Adam optimizer \cite{Adam} and the L1 loss function for training with a batch size of 32 for 1000 epochs based on the cosine annealing algorithm. And we localize the initial learning rate to $1 \times 10^{-5}$, and perform 50 rounds of warm-up to increase the learning rate to $1 \times 10^{-4}$ before training starts. 
We employ re-parameterization and cumulative training strategy to enhance the generalization performance of model across all scaling factors.

\subsection{Comparison with state-of-the-art methods}
\subsubsection{Quantitative analysis} We first compare the proposed FIT with other SOTA methods on the DIV2K validation set \cite{div2k} with EDSR \cite{edsr} and RDN \cite{rdn} as encoders, and the specific results are shown in Table \ref{sota_div}. Our network achieves the best results at all magnifications. In addition, we compare the results on other commonly used test datasets as shown in Table \ref{sota_test} with RDN \cite{rdn} as an encoder, Our method demonstrates remarkable performance improvements across all datasets.
%especially on Urban100 \cite{U100}, where the abundance of periodic structural patterns in the dataset can be effectively captured through frequency information. All the results indicates that FIT outperforms existing SOTA methods.

\subsubsection{Qualitative analysis} We conducted a series of qualitative experiments using RDN as an encoder on the DIV2K validation set \cite{div2k}, Urban100 \cite{U100} and Set14 \cite{set14} as shown in Fig. \ref{integer}. In the Monarch \cite{set14}, it can be seen that the image enhanced by LIIF shows marked blurring and texture errors. The other methods have improved by introducing different additional information, but artifacts remain observable, especially in the flowers at the center and in the grain on the left side of the Monarch. FIT got the best enhancement by introducing and utilizing frequency information. In the DIV2K validation set 0825 \cite{div2k}, FIT reconstruct the clearest images, especially the letters "M", "R", and "A". In the DIV2K validation set 0847 \cite{div2k}, the content of license plate reconstructed by the other methods is not clear. FIT can produce the clear edges of numbers. In the Urban100 0005 \cite{U100}, FIT reconstructs the clearest texture of the building image, and in particular the windows at the top of the image are recognizable.

Fig. \ref{noninteger} shows the results of different methods for super-resolution of progressively increasing non-integer scales of images when using RDN \cite{rdn} as an encoder. We amplified the text image using predetermined non-integer multiplication factors ×1.8, ×2.5, ×3.3 and ×4.2. Compared to the images amplified by other models, our model clearly reproduces the words “Yasashii” and “Akuma” in the first line of the image, and the words “Love” and “Letter” in the fourth line of the image are identifiable. 

\renewcommand{\arraystretch}{1.5}
\begin{table*}[t]
	\tiny
	%\scriptsize
	\centering
	\resizebox{0.9\linewidth}{!}{
		\begin{tabu}{l|ccccc|ccccc}
			\hline
			\specialrule{0em}{1pt}{0pt}
			\multirow{2}{*}{Method}
			&\multicolumn{5}{c|}{Set5 
				\cite{set5} 
			}  &\multicolumn{5}{c}{Set14 
				\cite{set14} 
			}    \\ 
			
			&$\times2$&$\times3$&$\times4$&$\times6$&$\times8$&$\times2$&$\times3$&$\times4$&$\times6$&$\times8$ \\
			
			\hline
			
			RDN 
			\cite{rdn}  
			&38.24&34.71&32.47&-&-&34.01&30.57&28.81&-&-                                                                            \\
			RDN-MetaSR 
			\cite{metasr}  
			&38.22 &34.63 &32.38 &29.04 &26.96 &33.98 &30.54 &28.78 &26.51&24.97            \\
			RDN-LIIF 
			\cite{liif}  
			&38.17&34.68 &32.50 &29.15  &27.14 &33.97 &30.53 &28.80&26.64 &25.15                          \\
			%RDN-UltraSR  &38.21 &34.67 &32.49 &29.33 &27.24 &33.97 & 30.59 &28.86 &26.69 &25.25       \\
			%RDN-ITSRN &38.23	&34.76	&32.55	&29.32	&27.25 &34.19	&30.59	&28.88	&26.68	&25.17  \\
			
			%RDN-IPE   &38.11 &34.68&32.51 &29.25&27.22 &33.94 &30.47 &28.75&26.58&25.09                      \\
			RDN-LTE 
			\cite{lte}  
			& 38.23&34.72&32.61 &29.32 &27.26 &34.09&30.58 &28.88 & 26.71 &25.16                           \\
			RDN-CLIT 
			\cite{clit}  
			&38.26 &34.80 &32.69 &29.39 &27.34 &34.21 &30.66 &28.98 &26.83 &25.35                                   \\
			RDN-SRNO 
			\cite{srno}
			&38.32 &34.84 &32.69 &29.38 &27.28 &34.27 &30.71 &28.97 &26.76 &25.26 \\
			RDN-CiaoSR
			\cite{ciaosr}
			&38.29	&34.85	&32.66	&29.46	&27.36
			&34.22	&30.65	&28.93	&26.79	&25.28
			\\
			RDN-AMI 
			\cite{amiassr}      
			&38.27	&34.80	&32.63	&29.43	&27.40
			&34.32	&30.78	&29.00	&26.82	&25.42
			\\
			%RDN-LMI    &38.30	&34.85	&32.72	&29.47	&27.41 &34.34	&30.80	&29.02	&26.83	&25.43
			%\\
			RDM-FIT(Ours)  &$\textbf{38.33}$	&$\textbf{34.87}$	&$\textbf{32.77}$	&$\textbf{29.49}$	&$\textbf{27.42}$ &$\textbf{34.47}$	&$\textbf{30.89}$	&$\textbf{29.04}$ &$\textbf{26.89}$	&$\textbf{25.45}$
			
			\\
			
			\specialrule{0em}{1pt}{0pt}
			\cline{2-11}
			%\hline
			\specialrule{0em}{1pt}{0pt}
			
			\multirow{2}{*}{ }     &\multicolumn{5}{c|}{BSD100 
				\cite{BSD100} 
			}  &\multicolumn{5}{c}{Urban100 
				\cite{U100} 
			}    \\ 
			&$\times2$&$\times3$&$\times4$&$\times6$&$\times8$&$\times2$&$\times3$&$\times4$&$\times6$&$\times8$ \\
			
			\specialrule{0em}{1pt}{0pt}
			\cline{2-11}
			%\hline
			\specialrule{0em}{1pt}{0pt}
			
			RDN 
			\cite{rdn}  
			& 32.34&29.26&27.72&-&-&32.89& 28.80&26.61&-&-                                                                            \\
			RDN-MetaSR 
			\cite{metasr} 
			& 32.33 &29.26 &27.71 &25.90 &24.83  &32.92&28.82 &26.55 &23.99&22.59           \\
			RDN-LIIF 
			\cite{liif}  
			&32.32&29.26&27.74&25.98&24.91 &32.87&28.82&26.68&24.20&22.79                              \\
			%RDN-UltraSR  &32.35&29.29&27.77& 26.01&24.96 &32.97&28.92& 26.78& 24.30 &22.87            \\
			%RDN-ITSRN &32.38	&29.32	&27.79	&26.01	&24.93 &33.07	&28.96	&26.77	&24.23	&22.81  \\
			%RDN-IPE  &32.31&29.28& 27.76& 26.00&24.93&32.97&28.82&26.76&24.26&22.87                      \\
			RDN-LTE 
			\cite{lte}  
			&32.36 &29.30 &27.77 &26.01 &24.95 &33.04 &28.97&26.81 &24.28 &22.88                          \\
			RDN-CLIT 
			\cite{clit}  
			&32.39 &29.34& 27.82 &26.07 & 25.00 &33.13 &  29.04 &26.91 &24.43 & 23.03                                \\
			RDN-SRNO 
			\cite{srno} 
			&32.43 &29.37 &27.83 &26.04 &24.99 &33.33 &29.14 &26.98 &24.43 &23.02 \\
			RDN-CiaoSR 
			\cite{ciaosr}
			&32.41	&29.34	&27.83	&26.07	&25.00 &33.30	&29.17	&27.11	&24.58	&23.13                             \\
			RDN-AMI 
			\cite{amiassr}
			&32.40	&29.36	&27.83	&26.06	&25.00 &33.31	&29.14	&27.03	&24.50	&23.16 \\
			%RDN-LMI  &32.44	&29.37	&27.83	&26.08	&25.02 &33.28	&29.20	&27.14	&24.62	&23.16
			%\\
			RDN-FIT(Ours)  &$\textbf{32.49}$	&$\textbf{29.40}$	&$\textbf{27.89}$	&$\textbf{26.13}$	&$\textbf{25.06}$
			&$\textbf{33.63}$	&$\textbf{29.46}$	&$\textbf{27.29}$	&$\textbf{24.74}$	&$\textbf{23.27}$
			\\
			\specialrule{0em}{1pt}{0pt}
			\hline
			
		\end{tabu}
	}
	
	\caption{Quantitative comparison with the SOTA methods on the benchmark test sets. The best results are shown in \textbf{Bold}.}
	\label{sota_test}
\end{table*}

\subsection{Ablation studies}In this section, we design a series of ablation experiments to investigate the role of each module. All ablation experiments were tested on the DIV2K validation set \cite{div2k} using EDSR \cite{edsr} as an encoder with a batch size of 16. The rest of the implementation details are consistent with the above.
\begin{table}[t]
	\tiny
	\setlength{\tabcolsep}{3.5mm}
	\centering
	\begin{tabu}{|c|c|ccccc|}
		\hline
		\multicolumn{1}{|c|}{\multirow{2}{*}{Module}} &{\multirow{2}{*}{Params}}& \multicolumn{4}{c|}{DIV2K val 100}                                              \\  
		\multicolumn{1}{|c|}{}    & \multicolumn{1}{c|}{}                                 & \multicolumn{1}{c}{$\times2$} & \multicolumn{1}{c}{$\times4$} & \multicolumn{1}{c}{$\times6$} &  \multicolumn{1}{c|}{$\times12$} \\ 
		\hline
		\specialrule{0em}{1pt}{0pt} 
		\hline
		Spatial               & \multicolumn{1}{l|}{$\textbf{ 6.6M}$} & \multicolumn{1}{l|}{{34.81}}          & \multicolumn{1}{l|}{29.12}          & \multicolumn{1}{l|}{26.84}          &    \multicolumn{1}{l|}{23.76}    \\ \hline
		FDF \cite{fadformer}                & \multicolumn{1}{l|}{{ 6.8M}}& \multicolumn{1}{l|}{{34.82}}          & \multicolumn{1}{l|}{29.13}          & \multicolumn{1}{l|}{26.84}          &    \multicolumn{1}{l|}{23.77}       \\ \hline
		FRB \cite{sfmnet}                & \multicolumn{1}{l|}{{ 6.8M}}& \multicolumn{1}{l|}{{34.82}}          & \multicolumn{1}{l|}{29.14}          & \multicolumn{1}{l|}{26.85}          &    \multicolumn{1}{l|}{23.77}       \\ \hline
		FIM                & \multicolumn{1}{l|}{{ 6.8M}}& \multicolumn{1}{l|}{$\textbf{34.84}$}          & \multicolumn{1}{l|}{$\textbf{29.16}$}          & \multicolumn{1}{l|}{$\textbf{26.87}$}          &   \multicolumn{1}{l|}{$\textbf{23.79}$}       \\ \hline
	\end{tabu}
	\caption{PSNR (dB) results for different convolution. The best performing results are highlighted in $\textbf{Bold}$.}
	\label{FIM_test}
\end{table}

\begin{table*}[t]
	\tiny
	\centering
	
	\resizebox{0.9\linewidth}{!}{
		\begin{tabu}{cc|ccccc|cc|cccc}
			\cline{1-13} 
			
			%& \multicolumn{5}{|c}{Num of convolutions} 
			%& \multicolumn{2}{c}{FEM}
			\multicolumn{2}{|c|}{FCSA}  
			& \multicolumn{5}{c|}{Num of subspaces in IISA} 
				& \multicolumn{2}{c}{Params}
			& \multicolumn{4}{|c|}{{DIV2K Val 100}}\\
			
			%%\multicolumn{2}{|c|}{}  
			%%& \multicolumn{5}{c|}{Num of subspace} 
			%%& \multicolumn{4}{c|}{}\\

			%& \multicolumn{1}{|c}{1}                        
			%& \multicolumn{1}{c}{2}                        
			%& \multicolumn{1}{c}{4}                        
			%& \multicolumn{1}{c}{8}                        
			%& 16                 
			%& \multicolumn{1}{c}{W}                         
			%& W/O                      
			\multicolumn{1}{|c}{W}                         
			& \multicolumn{1}{c|}{W/O}       
			& \multicolumn{1}{c}{0}                        
			& \multicolumn{1}{c}{2}                        
			& \multicolumn{1}{c}{4}                         
			& \multicolumn{1}{c}{8}                        
			& \multicolumn{1}{c|}{16}
			& \multicolumn{1}{c}{}
			& \multicolumn{1}{c}{}                    
			& \multicolumn{1}{|c}{$\times 2$}
			& \multicolumn{1}{c}{$\times 4$}               
			& \multicolumn{1}{c}{$\times 6$}               
			& \multicolumn{1}{c|}{$\times12$}                                                   \\ 
			\cline{1-13} 
			\specialrule{0em}{1pt}{0pt}
			\cline{1-13}
			\specialrule{0em}{1pt}{0pt}
			
			%& \multicolumn{1}{|c}{ \checkmark }                        
			%& \multicolumn{1}{c}{}                         
			%& \multicolumn{1}{c}{}                         
			%& \multicolumn{1}{c}{}                         
			%&              
			%& \multicolumn{1}{|c}{ \checkmark }                         %&                          
			\multicolumn{1}{|c}{}                         
			& \multicolumn{1}{c|}{\checkmark}              
			& \multicolumn{1}{c}{\checkmark}                         
			& \multicolumn{1}{c}{}                         
			& \multicolumn{1}{c}{}                         
			& \multicolumn{1}{c}{}                         
			&  \multicolumn{1}{c|}{}  
			&  \multicolumn{2}{c}{$\textbf{6.8M}$}                       
			&  \multicolumn{1}{|c|}{34.84}        
			& \multicolumn{1}{c|}{29.16}                         
			& \multicolumn{1}{c|}{26.87}                         
			&\multicolumn{1}{c|}{23.79}                                                         \\ 
			\cline{1-13} 
			
			\multicolumn{1}{|c}{}                         
			& \multicolumn{1}{c|}{\checkmark}              
			& \multicolumn{1}{c}{}                         
			& \multicolumn{1}{c}{\checkmark}                         
			& \multicolumn{1}{c}{}                         
			& \multicolumn{1}{c}{}                         
			&  \multicolumn{1}{c|}{}
			&  \multicolumn{2}{c}{$\textbf{6.8M}$}                          
			&  \multicolumn{1}{|c|}{34.89}        
			& \multicolumn{1}{c|}{29.19}                         
			& \multicolumn{1}{c|}{26.91}                         
			&\multicolumn{1}{c|}{23.83}                                                         \\ 
			\cline{1-13}
			
			\multicolumn{1}{|c}{}                         
			& \multicolumn{1}{c|}{\checkmark}              
			& \multicolumn{1}{c}{}                         
			& \multicolumn{1}{c}{}                         
			& \multicolumn{1}{c}{$\large{\ast}$}                         
			& \multicolumn{1}{c}{}                         
			&  \multicolumn{1}{c|}{}
			&  \multicolumn{2}{c}{6.9M}                          
			&  \multicolumn{1}{|c|}{34.85}        
			& \multicolumn{1}{c|}{29.17}                         
			& \multicolumn{1}{c|}{26.88}                         
			&\multicolumn{1}{c|}{23.80}                                                         \\ 
			\cline{1-13}  
			
			\multicolumn{1}{|c}{}                         
			& \multicolumn{1}{c|}{\checkmark}              
			& \multicolumn{1}{c}{}                         
			& \multicolumn{1}{c}{}                         
			& \multicolumn{1}{c}{\checkmark}                         
			& \multicolumn{1}{c}{}                         
			&  \multicolumn{1}{c|}{}
			&  \multicolumn{2}{c}{6.9M}                          
			&  \multicolumn{1}{|c|}{34.89}        
			& \multicolumn{1}{c|}{29.20}                         
			& \multicolumn{1}{c|}{26.91}                         
			&\multicolumn{1}{c|}{23.83}                                                         \\ 
			\cline{1-13} 
			
			\multicolumn{1}{|c}{\checkmark}                         
			& \multicolumn{1}{c|}{}              
			& \multicolumn{1}{c}{}                         
			& \multicolumn{1}{c}{}                         
			& \multicolumn{1}{c}{\checkmark}                         
			& \multicolumn{1}{c}{}                         
			&  \multicolumn{1}{c|}{}
			&  \multicolumn{2}{c}{7.1M}                          
			&  \multicolumn{1}{|c|}{$\textbf{34.92}$}        
			& \multicolumn{1}{c|}{$\textbf{29.23}$}                         
			& \multicolumn{1}{c|}{$\textbf{26.94}$}                         
			&\multicolumn{1}{c|}{$\textbf{23.85}$}                                                         \\ 
			\cline{1-13} 
			
			\multicolumn{1}{|c}{}                         
			& \multicolumn{1}{c|}{\checkmark}              
			& \multicolumn{1}{c}{}                         
			& \multicolumn{1}{c}{}                         
			& \multicolumn{1}{c}{}                         
			& \multicolumn{1}{c}{\checkmark}                         
			&  \multicolumn{1}{c|}{}
			&  \multicolumn{2}{c}{7.1M}                          
			&  \multicolumn{1}{|c|}{34.88}        
			& \multicolumn{1}{c|}{29.19}                         
			& \multicolumn{1}{c|}{26.89}                         
			&\multicolumn{1}{c|}{23.81}                                                         \\ 
			\cline{1-13} 
			
			\multicolumn{1}{|c}{}                         
			& \multicolumn{1}{c|}{\checkmark}              
			& \multicolumn{1}{c}{}                         
			& \multicolumn{1}{c}{}                         
			& \multicolumn{1}{c}{}                         
			& \multicolumn{1}{c}{}                         
			&  \multicolumn{1}{c|}{\checkmark}
			&  \multicolumn{2}{c}{7.7M}                          
			&  \multicolumn{1}{|c|}{34.84}        
			& \multicolumn{1}{c|}{29.17}                         
			& \multicolumn{1}{c|}{26.87}                         
			&\multicolumn{1}{c|}{23.80}                                                         \\ 
			\cline{1-13}

	\end{tabu}}
	
	\caption{The module ablation experiments conducted on FUSAM on the DIV2K validation set, where "$\ast$" indicates that all subspaces focus on the spatial domain. The best performing results are highlighted in $\textbf{Bold}$.}
	\label{DFSAM_test}
\end{table*}

\subsubsection{Effectiveness of FIM}We use FIM to incorporate frequency information to the model. 
%in order to validate the effectiveness of FIM. We replace our module with other frequency information extraction modules with conventional spatial information extraction module for ablation experiments.
%Table \ref{FIM_test} shows the outcome of our quantitative comparison of the effects of different modules, and FIM achieved the best results at all magnifications.
We replace the FIM with spatial module, Frequency-Domain Fusion (FDF) from FADformer \cite{fadformer} and Frequency Block (FRB) from SFMNet \cite{sfmnet} to verify the performance of the different modules. 
Table \ref{FIM_test} shows FIM achieved the best results at all magnifications.
Visual feature maps are used to visually analyze the effectiveness of different modules in the introduction of frequency information in Fig. \ref{visual}. Visual feature map obtained by the spatial module is blurriest, indicating the lack of frequency information significantly deteriorates the detail characterization. 
%Although the visual feature map generated by FDF \cite{fadformer} enhance the edge characterization it is still unsatisfactory, suggesting that FDF losses frequency information due to destroying the channel dependency.
%The visual feature maps of FRB\cite{sfmnet} enhance focus to the background while ignoring details, owing to the fact that FRB inevitably transforms detail into non-extractable non-numbers such as nan, posinf and neginf.
%The visualized feature maps of FDF \cite{fadformer} and FRB \cite{sfmnet} are suboptimal because of the loss of frequency information due to the destruction of channel dependence and the loss of unextractable details in polar coordinate.
Clearer visualized feature maps of FDF and FRB show that the introduction of frequency information effectively improves the detail characterization. But the blurring and artifacts are still significant due to the reorganization of the frequency information in the channel dimension by FDF and the transformation of the frequency information into the polar coordinate by FRB inevitably lead to the loss of frequency information.
The clearest visual feature map captured by FIM, demonstrating the detail characterization is fully enriched through the lossless introduction of frequency information.
%The clarity of the visualized feature maps produced by FIM is superior to existing methods demonstrating its superiority in introducing frequency information.
%FRB \cite{sfmnet} remaps complex tensors into magnitude tensors and phase tensors, but some tensors representing details are incorrectly mapped as non-numbers, causing FRB \cite{sfmnet} to focus on the background and ignore the details. FIM obtains the clearest visual feature map due to the lossless extraction of frequency information.

\subsubsection{Effectiveness of IISA}
\begin{figure}[t]
	\centering
	\begin{subfigure}{1\linewidth}
		\centering
		\includegraphics[width=1\linewidth]{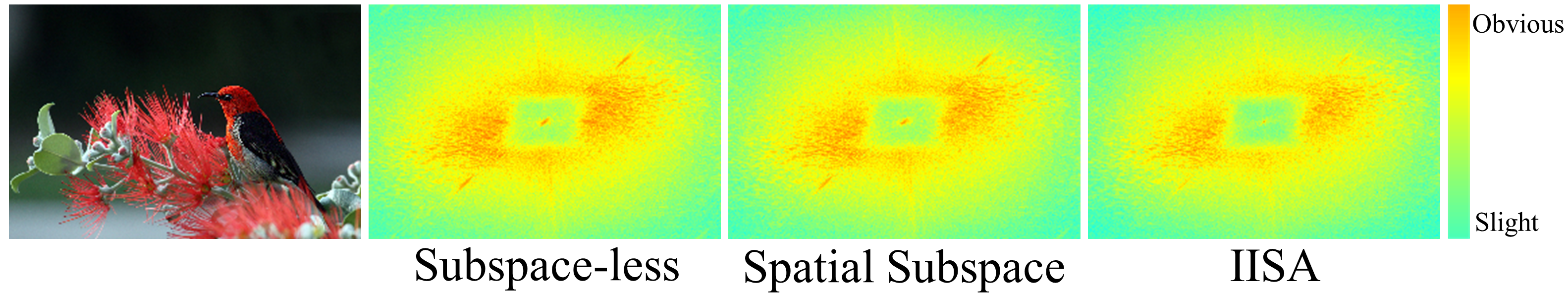}
	\end{subfigure}
	\caption{Visualization experiments on the effect of subspace and cross-domain information interactions on frequency errors. }
	\label{pinpu}%文中引用该图片代号
\end{figure}

IISA alternately projects spatial and frequency information into subspace to exploit the spatial-frequency interrelationship for cross-domain synergy of information. Table \ref{DFSAM_test} shows the ablation experiments we performed for IISA regarding the number of subspaces and types of information interaction, IISA achieves the best results at all  magnifications for cross-domain information interaction in $4$ subspaces.
%We also refer to the mean error map \cite{mem} to design a frequency error map (FEM) as shown in Fig.  to qualitatively analyse the effect of subspace and cross-domain information interaction on effect of ASSR. 
We refer to the mean error map \cite{mem} to design the frequency error map shown in Fig. \ref{pinpu} to analyze frequency fidelity of subspace-less module, spatial subspace module and IISA in the frequency domain.
%We show the effect of subspace and cross-domain information interactions in the frequency domain in Fig. \ref{pinpu}.
In these maps, from center to edge represents the frequency changing from low to high, and the color changing from red to green represents the error from obvious to slight.
%The frequency error of the FEM located in the center is more slighter compared to the one on the left, indicating the module achieves a simple information interaction through multiple-subspace, but the results are suboptimal. 
The frequency error of the spatial subspace module is slight compared to subspace-less module, indicating the spatial information interaction through multiple-subspace can increase the frequency fidelity in a way. IISA exhibits the slightest frequency error demonstrating the excellent synergy of spatial and frequency information increases frequency fidelity through the cross-domain information interaction in multiple-subspaces.
%The FEM on the left indicates that IISA achieved the most obvious error in all frequency bands without subspace. When IISA project the multi-space without the cross-domain information interaction, the error is slighter but sub-optimal. IISA achieves the slightest frequency error in projecting spatial and frequency information alternately to subspace for cross domain information interaction.

\subsubsection{Effectiveness of FCSA}
\begin{figure}[h]
	\centering
	\begin{subfigure}{1\linewidth}
		\centering
		\includegraphics[width=1\linewidth]{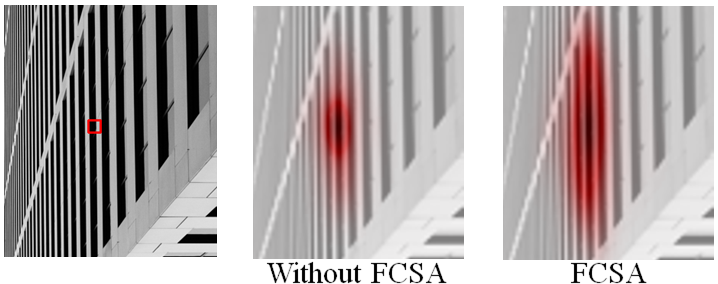}
	\end{subfigure}
	\caption{Visualization experiments using LAM to demonstrate the effect of FCSA on capturing global context.}
	\label{lam}%文中引用该图片代号
\end{figure}

FCSA is designed to obtain global context. According to the Table \ref{DFSAM_test}, it can be seen that FCSA is indispensable at all magnifications. Moreover, we investigated the impact of FCSA on the receptive field using local attribute mapping (LAM) \cite{lam} as shown in Fig. \ref{lam}. In these maps, the red region are the context region exploited by the model. Higher saturation of the red color represents greate capitalization of the region. The context region of the right LAM is significantly larger than the left in terms of coverage area and color saturation, proving that FCSA can significantly improve the ability of the network to capture global context.

\subsection{Limitations}FIT outperforms existing methods by integrating and utilizing frequency information. But FIT can be further improved in adaptive exploitation of frequency information.
%But FIT can be further optimized in terms of performance and computational resource. 
First, we can dynamically adjust the exploitation frequency information according to the magnification. Secondly, adopting location coding applicable to frequency information rather than using existing spatial information coding can further enhance the effectiveness in utilizing frequency information. 
%In addition, assigning different weights according to the influence of different frequency bands can save computational resources. 
In addition, weighted exploitation of information from different frequency bands.
These deserve further study and are important directions for future exploration.

	\section{Conclusion}In this work, we propose the Frequency-Integrated Implicit Transformer (FIT) for Arbitrary-Scale Single Image Super-Resolution (ASSR) by lossless introduction and efficient utilization of frequency information. FIT is consists of Frequency Incorporation Module (FIM), and Frequency Utilization Self-Attention Module (FUSAM).
	FIM realizes lossless incorporation of frequency information through FFT and real-imaginary mapping. In FUSAM, IISA utilizes spatial-frequency interrelationships for cross-domain interaction of information, while FCSA leverages the advantage of Self-Attention (SA) in acquiring context to efficiently capture the global nature of frequency.
	%Numerous experiments have demonstrated that the high definition images attained by FIT through the introduction and utilization of frequency information at all magnifications are competitive with those obtained by a number of state-of-the-art (SOTA) methods.
	Numerous experiments demonstrated FIT can obtain high-resolution images at all magnifications superior to existing methods.  
	Visual feature map show FIM in enriching detail characterization through lossless introduction of frequency information. Frequency error map (FEM) demonstrates IISA synergizes spatial and frequency information through subspace projection and information cross-domain interaction to increase frequency fidelity. LAM proves the effectiveness of FCSA to capture global context.
	%FIT is a network that enhances the ASSR effect through lossless introduction and efficient utilization of frequency information and 
	FIT promotes the introduction and utilization of frequency information in ASSR and is expected to be applied in image denoising, image deblurring, and super resolution.
	Future work could focus on the adaptive use of frequency information.
	%Future performance will focus on exploring the dynamics of frequency information under different magnifications and on designing frequency position coding, while conserving computational resources by assigning weights to different frequency bands.
	%Future research could focus on exploring the dynamic changes of frequency information under different scaling factors to improve the generalization performance of model at various magnifications, assigning different weights to information from different frequency bands to allocate computational resources efficiently, and performing position encoding based on the distribution patterns of frequency information to integrate positional information into the frequency domain. 
	
	{\small
		\bibliographystyle{unsrt}
		\bibliography{egbib.bib}
	}
	
	\vspace{-10mm}
	
	\begin{IEEEbiography}[{\includegraphics[width=1in,height=1.25in,clip,keepaspectratio]{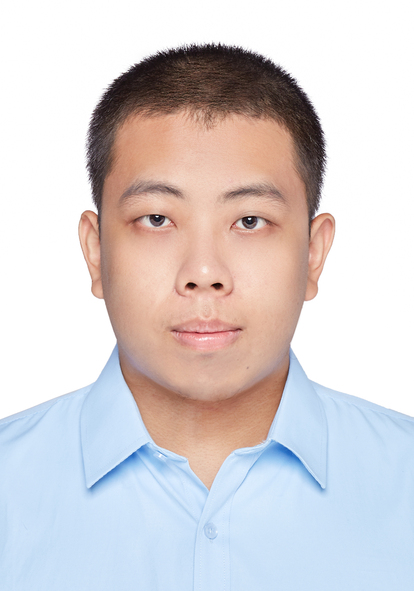}}]{Xufei Wang}
		received the B. S. degree from QingDao University in 2022, and will receive the M.S. degree in June 2026 from Anhui University, 
		School of Electronic and Information Engineering, majoring in Information and 
		Communication Engineering. He mainly works on low-level computer vision tasks, 
		including image denoising, image super-resolution.
	\end{IEEEbiography}

	\vspace{-10mm}
	
	\begin{IEEEbiography}[{\includegraphics[width=1in,height=1.25in,clip,keepaspectratio]{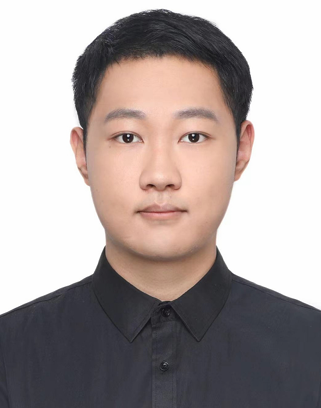}}]{Fei Ge}
		will receive the M.S. degree in June 2026 from Anhui University, 
		School of Electronic and Information Engineering, majoring in Electronic Information. He mainly works on low-level computer vision tasks, 
		including image deblurring, image super-resolution.
	\end{IEEEbiography}
	
		\vspace{-10mm}

	\begin{IEEEbiography}[{\includegraphics[width=1in,height=1.25in,clip,keepaspectratio]{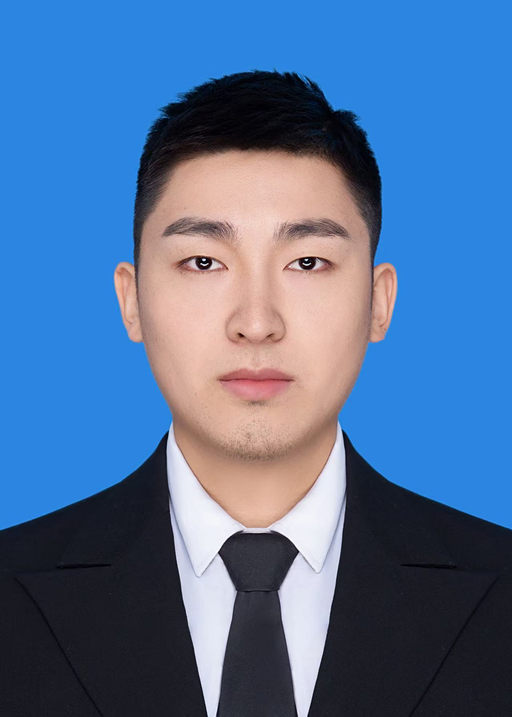}}]{Jinchen Zhu}
		will receive the M.S. degree in June 2025 from Anhui University, School of Electronic and Information Engineering, majoring in Information and Communication Engineering. He mainly works on low-level computer vision tasks, including image denoising, image super-resolution.
	\end{IEEEbiography}
	
	\vspace{-10mm}
	
	\begin{IEEEbiography}[{\includegraphics[width=1in,height=1.25in,clip,keepaspectratio]{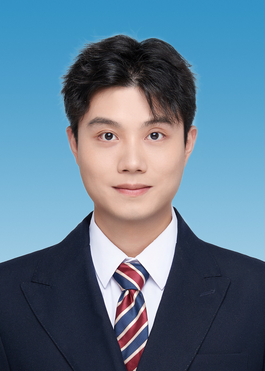}}]{Mingjian Zhang}
		 received the B. S. degree from Hunan University of Science and Technology in 2018, and will gain M.S. degree in 2025 from Anhui University, School of Electronic and Information Engineering, majoring in Information and Communication Engineering. He mainly works on image super-resolution.
	\end{IEEEbiography}
	
	\vspace{-10mm}
	
	\begin{IEEEbiography}[{\includegraphics[width=1in,height=1.25in,clip,keepaspectratio]{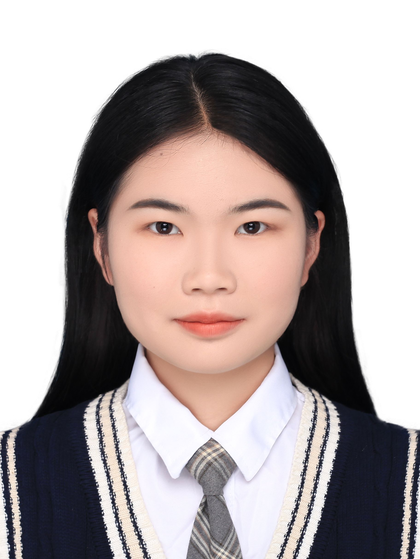}}]
	{Qi Wu}
	  	received the B. S. degree from Henan University in 2024, and will gain M.S. degree in 2027 from Anhui University, School of Electronic and Information Engineering, majoring in New-Generation Electronic Information Technology. She mainly works on image super-resolution, image blurring.
	\end{IEEEbiography}
	
	\vspace{-120mm}
	
	\begin{IEEEbiography}[{\includegraphics[width=1in,height=1.25in,clip,keepaspectratio]{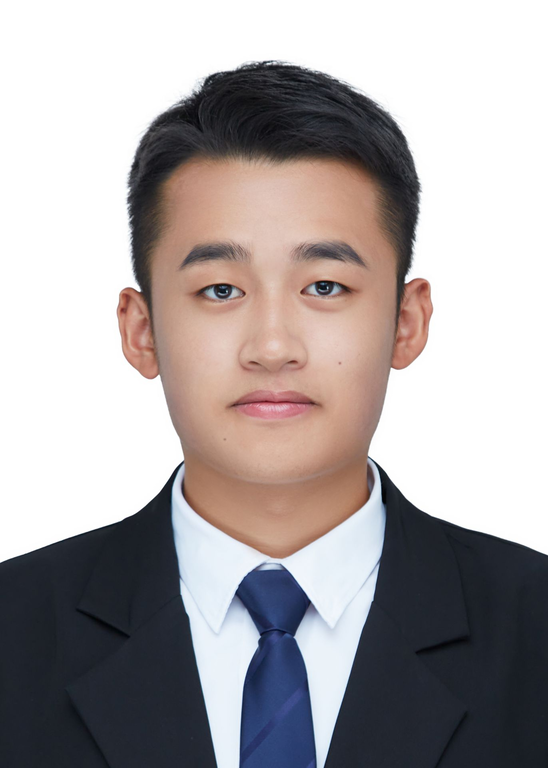}}]
		{Jifeng Ren}
		 will receive the B. S. degree from Anhui University in 2027, majoring in Electronic science and technology. He mainly works on image super-resolution.
	\end{IEEEbiography}
	
	\vspace{-120mm}
	
	\begin{IEEEbiography}[{\includegraphics[width=1in,height=1.25in,clip,keepaspectratio]{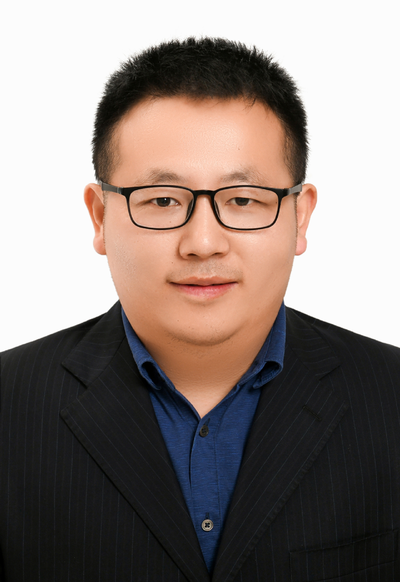}}]{Shizhuang Weng}
		received his PhD from University of Science and Technology of 
		China, is an associate professor in the School of Electronic Information 
		Engineering of Anhui University, and a member of the Chinese Society of 
		Artificial Intelligence. He is engaged in the research of computer vision, image 
		processing and deep learning applications. He has presided over and participated 
		in the National Natural Science Foundation of China, Anhui Key Research and 
		Development Program, Provincial Natural Science Foundation, Provincial 
		Science and Technology Police Project, Provincial Natural Science Research 
		Project of Education Department, National Natural Science Foundation of China, National Science 
		and Technology Support and Enterprise Entrusted Development Project. Research results won the 
		second prize of Anhui Province Electronic Information Science and Technology. He is an expert in 
		the evaluation of scientific and technological projects in Anhui, Zhejiang and Jiangxi Province.
	\end{IEEEbiography}
	%\input{texts4neurips/supp}
	%\bibliographystyle{plainnat}
	%\bibliography{ref,unsrtnat}
	%\bibliographystyle{unsrtnat}

\end{document}